\documentclass[letterpaper, 10 pt, conference]{ieeeconf}  % Comment this line out if you need a4paper

\IEEEoverridecommandlockouts                              % This command is only needed if 
\title{\LARGE \bf
An MCTS-DRL Based Obstacle and Occlusion Avoidance Methodology in Robotic Follow-Ahead Applications
}

\author{Sahar Leisiazar$^{1}$, 
Edward J. Park$^{1}$,
Angelica Lim$^{2}$ and
Mo Chen$^{2}$ % <-this % stops a space
\thanks{*This work was supported by the Huawei-SFU Visual Computing Joint Lab.}% <-this % stops a space
\thanks{$^{1}$School of Mechatronics Systems Engineering, Simon Fraser University, Canada
        {\tt\small \{sleisiaz,ed\_park\}@sfu.ca}}%
\thanks{$^{2}$School of Computing Sciences, Simon Fraser University, Canada
        {\tt\small \{angelica, mochen\}@sfu.ca}}%
}

\usepackage{subcaption}
\usepackage{graphicx}
\usepackage{amsmath}
\usepackage{algorithm2e}

\usepackage{tabularx}

\usepackage{multirow}
\usepackage[table,xcdraw]{xcolor}
\usepackage{tabularray}
\usepackage{float}
\usepackage{graphicx}
\usepackage[export]{adjustbox}

\usepackage{siunitx}
\usepackage{booktabs}

\usepackage{cite}

\usepackage{todonotes}

\begin{document}

\maketitle
\thispagestyle{empty}
\pagestyle{empty}

%%%%%%%%%%%%%%%%%%%%%%%%%%%%%%%%%%%%%%%%%%%%%%%%%%%%%%%%%%%%%%%%%%%%%%%%%%%%%%%%
\begin{abstract}

We propose a novel methodology for robotic follow-ahead applications that address the critical challenge of obstacle and occlusion avoidance. Our approach effectively navigates the robot while ensuring avoidance of collisions and occlusions caused by surrounding objects. To achieve this, we developed a high-level decision-making algorithm that generates short-term navigational goals for the mobile robot. Monte Carlo Tree Search is integrated with a Deep Reinforcement Learning method to enhance the performance of the decision-making process and generate more reliable navigational goals. 
Through extensive experimentation and analysis, we demonstrate the effectiveness and superiority of our proposed approach in comparison to the existing follow-ahead human-following robotic methods.
Our code is available at https://github.com/saharLeisiazar/follow-ahead-ros.

\end{abstract}

%%%%%%%%%%%%%%%%%%%%%%%%%%%%%%%%%%%%%%%%%%%%%%%%%%%%%%%%%%%%%%%%%%%%%%%%%%%%%%%%
\section{INTRODUCTION}

Human-robot interaction involves robots performing stable, real-time, and safe interactions with a target person in same natural space. 
In particular, human-following robots (HFR) maintain a consistent distance and orientation between the target person and the robot. 
The ability to follow a person not just behind, but in front or side by side, in complex environments is important for applications such as autonomous suitcases or shopping carts, search and rescue robots during emergencies, capturing the video of physical activities, guiding visitors in an airport or hospitals or monitoring elderly people.

There are three types of HFRs: those that follow from behind, those that follow side-by-side, and those that follow-ahead of a person.
In recent years, there has been a significant amount of research focused on following from behind or side-by-side \cite{c4, c6, c10}.
Following \textit{ahead} of a person is a more intricate task for a robot, as it necessitates the ability to predict the person's intention and trajectory in advance, and to navigate ahead of the user. 
Previous works have proposed various methods to address this challenge.
Moustris et al. \cite{c7} proposed a front-following robot that uses a user intention recognition algorithm. 
Nikdel et al. \cite{c8} employed Deep Reinforcement Learning (DRL) to estimate the human trajectory and used classical trajectory planning methods to generate short-term navigational goals to guide a follow-ahead robot. 
Mahdavian et al. \cite{c11} trained a non-autoregressive transformer model to predict human intention for follow-ahead robotic applications.

\begin{figure}
\centering
\begin{subfigure}[b]{0.49\columnwidth}
\includegraphics[width=\columnwidth]{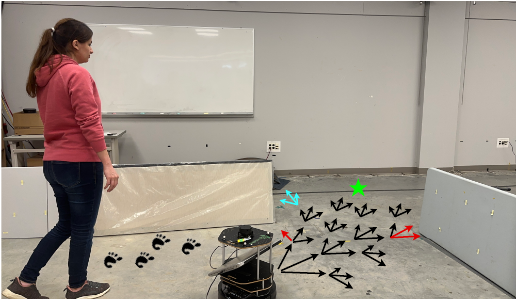}
\end{subfigure}
\begin{subfigure}[b]{0.49\columnwidth}
\includegraphics[width=\columnwidth]{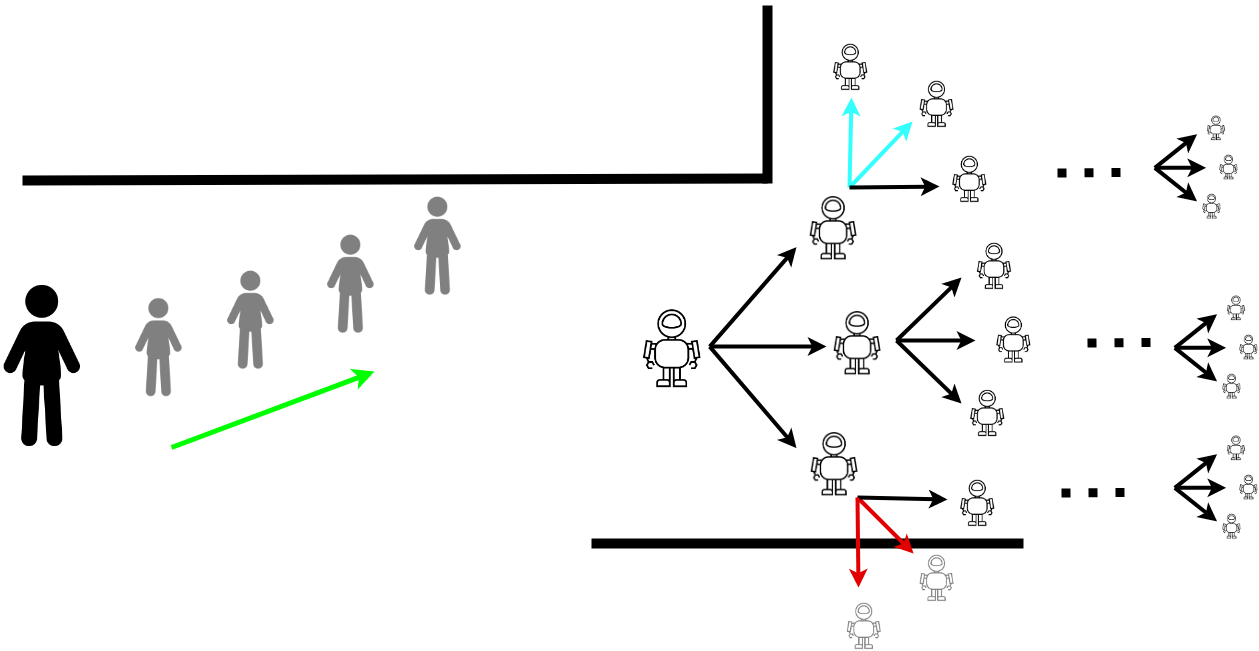}
\end{subfigure}
\caption{Illustration of the MCTS-DRL framework utilizing human future estimation for goal generation. The search tree is expanded to identify the best goal point to follow ahead of the person, while avoiding occlusion and collision. The resulting goal point is indicated by a green star, while red and blue arrows represent paths leading to collision and occlusion, respectively. The MCTS algorithm expands a tree to find the best navigational goal for the robot in order to follow-ahead of the target person and avoid collision and occlusion caused by surrounding objects.}
\label{fig:general}
\end{figure}

Previous research on follow-ahead robots have focused on occlusion-free environments; however, in real-world scenarios, a robot may encounter various forms of occlusions while following a person in complex environments. 
Several works addressed this challenge in follow behind or side-by-side scenarios using various sensors including lidar sensors  and cameras, and with controllers such as PD and MPC \cite{c1, c2, c3, c5, c12, c13}.

To the best of our knowledge, no existing follow-ahead HFR method addresses the problem of occlusion and obstacle avoidance. 
With that in mind, we aim to improve upon the idea of \cite{c8} and \cite{c11} by integrating Monte Carlo tree search (MCTS) with DRL to make high-level decisions to avoid undesirable situations such as occlusion and collision with obstacles. 
It may be extremely difficult or data inefficient to apply standard DRL, which trains an agent to make a sequence of low-level decisions at every time-step to perform a task.
MCTS is a powerful approach for designing game-playing agents and solving sequential decision-making problems. The method relies on an intelligent tree search that balances exploration and exploitation \cite{c9,c14}. 
In robotics, Dam et al. \cite{c15} utilized MCTS for path planning of a robotic arm in partially observable environments. The results demonstrated the effectiveness of MCTS in selecting optimal and feasible paths in fully and partially observable environments, respectively. 
MCTS-based path planning algorithm was also proposed by Qian et al. \cite{c16} for maximizing the average throughput of a Unmanned Aerial Vehicle (UAV).

In this paper, we propose a hierarchical decision-making architecture for a mobile robot, in which the robot must observe the target person's pose and the locations of surrounding obstacles relative to itself, and determine how to avoid collisions and occlusions caused by the obstacles while following ahead of the person. 
To achieve this, we propose integrating MCTS with DRL to make high-level decisions, specifically generating navigational goal points for the robot. 
This approach leaves the low-level execution, which is navigating toward the generated goal points, to be handled by well-established classical controllers. 
Such a hierarchical approach has been shown to be effective in improving data efficiency in recent works \cite{c8, c17}.
Our main contributions are as follows:
\begin{itemize}
    \item For the first time, we propose a novel methodology for the robotic follow-ahead application that addresses the challenge of obstacle and occlusion avoidance. Our method effectively navigates the robot in front of the target person while ensuring avoidance of collisions and occlusions caused by surrounding objects.
    \item We developed a high-level decision-making algorithm that generates navigational goals for the mobile robot, and adapted MCTS, which was originally designed for discrete systems, to a continuous setting, particularly in the context of path planning for mobile robots.
    \item We integrated MCTS with DRL to enhance the performance of the decision-making process and generate more consistent and reliable navigational goals in any complex environments with a variety of obstacles.
\end{itemize}

\section{background}
\subsection{Deep Reinforcement Learning}
DRL involves the use of a policy and a reward signal. The agent learns to optimize its actions to maximize the discounted sum of rewards, or \textit{return}, through a process of trial and error. 
The environment is modeled as a Markov Decision Process (MDP), where each state of the system is determined solely by the previous state and the action taken by the agent to transition to the next state.
In this paper, we employed the Double Deep Q-Network (DDQN) method \cite{c18} to train our agent. 

The reward function, which is inspired by \cite{c8} with certain modifications, is defined to ensure that the robot navigates in front of the target person and maintains a human-robot relative distance within the range of $[1,2]$ meters. 
The reward function is computed by \eqref{eq:r} and is composed of two parts: $r_a$ and $r_d$. 
The angle reward, $r_a$, guides the robot to remain in front of the person and the distance reward, $r_d$, encourages the robot to maintain the defined distance range in front of the person.
In the following equations, $\alpha$ is the angle between the person-robot vector and the person-heading vector, and $d_h$ represents the human-robot relative distance.

\begin{subequations}\label{eq:r}
\begin{align}
&r = \max(r_{d} + r_{\alpha}, -1), \\ 
&\quad \text{where } r_d = \left\{
    \begin{array}{ll}
        -(1-d_h) & 0.5 < d_h < 1  \\
        0 & 1 < d_h < 2 \\
        -0.25(d_h-1) & d_h < 5 \hspace{2mm} \text{ or } \hspace{2mm} d_h > 2 \\
        -1 & \text{otherwise}
    \end{array}
\right. \label{eq:rd} \\ 
&\quad \text{and } r_\alpha =  (45 - \alpha)/45 \label{eq:ra}
\end{align}
\end{subequations}

\subsection{Monte Carlo Tree Search}
MCTS analyzes the most promising high-level decisions made by sampling the search space to expand the tree search. 
The application of MCTS is based on many rollouts or simulated trajectories. 
The algorithm can be broken down into the following steps:
\begin{itemize}
    \item Select good child nodes, starting from the root node that represent system states toward a leaf node;
    \item Expand the leaf node and creating its child nodes;
    \item Run a simulated rollout from the created child node and compute its value; and
    \item Back propagating the simulated result and update the value of the parents nodes.
\end{itemize}

The selection of nodes is based on the Upper Confidence Bound (UCB) \cite{c19} with the node having the highest UCB being chosen: 

\begin{equation}
\label{eq:ucb}
    \text{UCB} = \frac{w}{n^c} + c\hspace{1mm} \sqrt{\frac{\log n^p}{n^c}}
\end{equation}

\noindent where $w$, $n^c$, and $n_p$ represent the value of the leaf node after a rollout, the number of visits to a node, and the number of visits to its parent node, respectively.
Eq. \eqref{eq:ucb} calculates the UCB value, which aids in balancing exploration and exploitation in the MCTS algorithm. A trade-off between the two factors is achieved by setting the parameter $c = 1.4$. 

\section{methodology}

In this paper, the proposed MCTS-DRL method makes high-level decisions and generates navigational goals for the robot. 
Then, the robot utilized the ROS navigation stack \cite{c20} and TEB planner \cite{c21} to navigate towards the goals.

\subsection{Observation Space, Action Space, and Robot Model}
We define the state of the system in \eqref{eq:S}, where $x,y,\theta$ correspond to the 2D pose in the global coordinate frame, while $r$ and $h$ refer to the robot and human, respectively. 
Each node of the MCTS represents this state, which contains the pose of both the robot and human; the application of MCTS will be explained in detail in Section \ref{sec:MCTS}.

\begin{equation}
\label{eq:S}
    \text{state} = (x_r, y_r, \theta_r, x_h, y_h, \theta_h)
\end{equation}

While the robot and person's global pose are used for the search tree, we employ the relative pose of the person with respect to the robot as the observation for the DRL model. The relative pose is defined as follows:

\begin{equation}
    \label{eq:obs}
    o = (x_p - x_r , y_p - y_r , \theta_p - \theta_r)
\end{equation}

To ensure compatibility with the MCTS, which operates on discrete action spaces, we define three distinct actions for each time step ($\delta t = 0.5$ s) in which the robot can go either straight, turn left or right for 45 degrees. 
Therefore, the robot's three possible next poses can be calculated as

\begin{equation}
\label{eq:nextS}
    \begin{bmatrix}
        x'_r \\ y'_r \\ \theta'_r
    \end{bmatrix} 
    = 
    \begin{bmatrix}
        x_r \\ y_r \\ \theta_r
    \end{bmatrix} 
    +
    \begin{bmatrix}
        d\cos(\theta + \psi) \\ d \sin(\theta + \psi) \\ \psi.
    \end{bmatrix}    
\end{equation}

\noindent where $\psi \in \{-45^\circ, 0^\circ , 45^\circ\}$ and by assuming the constant velocity of $V=0.6$ m/s, the robot travels $d = 0.3$ m at each time-step of $\delta t = 0.5$ s. 

\subsection{Sensors and Human Motion Model}

The MCTS-DRL method leverages both the current and predicted position and orientation (pose) of a target person. Depth data obtained from an RGB-D camera is utilized to measure and predict the person's future movement. To determine the global position of the surrounding obstacles, we utilized the occupancy map of the environment obtained from the SLAM-gmapping algorithm \cite{c23}.

We sample the human's positions to predict their future poses. This is done by fitting a line to the person's pose over a three-second period, and extrapolating the predicted pose $T=3$ seconds into the future based on this line.

In general, human trajectory prediction can be performed using other techniques proposed in recent literature \cite{c11, c22}; however, this was not the main focus of the paper.

\subsection{Deep Reinforcement Learning}

We first used DDQN to train, in an obstacle-free environment, to obtain a $Q$ function, which takes as input the observation $o$ calculated using \eqref{eq:obs}, and one of three robot actions $a_i$, and produces as output the estimated expected return, $R_i$:

\begin{equation}
    R_i = Q(o, a_i)
\end{equation}

Prior to training, we generated a dataset of ten million tuples $(o, a, R, o')$ in a 2D simulation environment. Each tuple contains the return ($R$) and subsequent observation ($o'$) resulting from a randomly chosen action ($a$). Human trajectories, including straight lines, turns, and wave-forms, were randomly assigned. The robot's initial pose relative to the human was also randomized. We applied a single action throughout each episode, computed the discounted sum of rewards as the return, and stored each tuple in a buffer. Eq. \ref{eq:return} is used to compute the return in which $\gamma = 0.99$ is the discounted factor, $n$ represents the length of the episode, and $r_i$ denoted the reward at each time-step.

\begin{equation}
\label{eq:return}
    R = \sum_{i=0} ^{n} \gamma^i r_i
\end{equation}

During the tree expansion of MCTS, the trained $Q(o, a_i)$ is utilized to estimate the expected reward of each node instead of employing random sampling which is explained in details in Section \ref{sec:MCTS}. 
It is noteworthy that the model was only trained in an obstacle-free environment due to the difficulty and data inefficiency of training a model that considers all obstacles in the environment while also accounting for the person's following behavior. Additionally, to enable the proposed algorithm to be applicable in any environment with diverse obstacles, obstacle and occlusion avoidances are only considered during the tree expansion phase.

\subsection{Monte Carlo Tree Search}
\label{sec:MCTS}

Following ahead of a target person presents a challenge due to its many possible trajectories and large state space. Finding the optimal action using DRL can be difficult, while MCTS can efficiently explore the decision space, simulate outcomes, and improve solution quality. 
Moreover, MCTS is a versatile algorithm that can be applied to a wide range of environments, including those with various obstacles, and it can effectively operate with any obstacle map.

The MCTS algorithm relies on random sampling to expand and evaluate its nodes, which can result in the generation of non-consistent trajectories for the robot. To address this limitation, we utilized the DRL model $Q(o, a_i)$ to help the process of evaluation and selection of leaf nodes during the tree expansion.

The process of tree expansion utilized in this study is described by Alg. \ref{al:mcts}. The algorithm takes in the current pose of the robot, as well as the current and predicted future pose of the human, along with the occupancy map of the environment as inputs. 
To adapt the proposed MCTS-DRL approach to a continuous setting, our method generates, every $\delta t = 0.5$ s, a navigational goal point for $T = 3$ s into the future in a receding horizon fashion.

At time $t_0$, the current pose of the human and robot are sampled, the state of the system is calculated using \eqref{eq:S} and considered as the root node of the search tree. The robot's next possible poses at time $t_0+\delta t$ are simulated with \eqref{eq:nextS} and through the predicted human pose, the algorithm calculates the system's next states and consider them as the child nodes of the root/parent node.
The process is repeated for each child node for the next time-steps to expand the tree as shown in Fig. \ref{fig:general}. Each branch/path can be expanded for the maximum number of six time-steps (totaling $T=3$ s) and since the size of the tree grows exponentially, we set a finite number of expansion for the purpose of real-time navigation.

\LinesNumbered
\RestyleAlgo{ruled}
\begin{algorithm}[t]
\caption{MCTS-DRL Approach}
\label{al:mcts}
\KwData{robot pose = $(x_r, y_r,\theta_r)^{t_0}$}
\KwData{$(x_h, y_h,\theta_h)^{t_0},...,(x_h, y_h,\theta_h)^{t_3}$}
\KwData{Occupancy map}
\KwResult{Navigational goal point}
\vspace{2mm}
parent node $= (x_{r}^{t_0}, y_{r}^{t_0}, \theta_{r}^{t_0}, x_{h}^{t_0}, y_{h}^{t_0}, \theta_{h}^{t_0}) $\;
\For{num of expansion}{
    \While{the parent node is not fully expanded} {
        child $\leftarrow$ simulate using an action\;
        \eIf{no collision} {
            \eIf{no occlusion } {
                % reward\_list $\leftarrow$ []\;
                % \While{human pose prediction available} {
                % $a \leftarrow \arg \max_a Q(o, a_i)$ \;\\
                $R \leftarrow  Q(o, a_i) $ \;
                child node state $\leftarrow$ simulate using the action \:
                % $o \leftarrow o'$ \;
                % insert $r$ into reward\_list\;
                % }
                % value $\leftarrow$ mean of rewards\_list\;
            } {
            value $\leftarrow -1$\;
            }
            parent value $\leftarrow$ parent value + value\;
        }{
        delete the child node\;
        }
    }
    parent node $\leftarrow$ A new leaf node with the highest UCB\;
}
goal point $\leftarrow$ leaf node with highest UCB ($c=0$)\;
\end{algorithm}

To ensure obstacle avoidance, we utilize the provided occupancy map of the environment. During the tree expansion process and the simulation of each child node, the algorithm verifies whether there are any obstacles in proximity to the node. If the child node was found to be located on or near any obstacles, the algorithm removes the node from the tree and terminates the tree expansion from that path. 
In Fig. \ref{fig:general}, the red arrows depict the paths to child nodes that collide with obstacles, while the blue arrows represent paths causing occlusion, both of which are subsequently removed from the tree.

The tree expansion process involves assigning values within the range of $[-1, 1]$ to the nodes. A value of 1 indicates that the robot remains in front of the human for the next $T =3$ s based on human trajectory estimation, while a value of $-1$ suggests that an occlusion or collision occurred or the robot is not following ahead of the human.
To assign these values, the algorithm first checks for any obstacles between the robot and human that may be causing occlusion or collision. If an obstacle is present, the assigned value is $-1$. However, if no obstacle is detected, the algorithm utilizes the trained DRL model to calculate the value of the child node.
The model is provided with the human-robot relative pose, $o$, obtained from the parent node and outputs the estimated expected return for each child node. It should be noted that the variable $w$ in \ref{eq:ucb} is equivalent to the $Q$ value acquired from the DRL model.
The value is then backpropagated through the parent nodes and used to update their values.

Upon expanding and simulating the child nodes of the parent node, the algorithm computes the UCB for each leaf node. The algorithm selects the node with the highest UCB and designates it as the new parent node. This process continues iteratively by expanding and simulating the children of the new parent node until the termination criterion is met.

After the completion of the tree expansion process, we evaluate the UCB value of all the leaf nodes in the tree with a constant value of $c=0$. The leaf node having the highest UCB value is then chosen. As previously mentioned, each node represents a potential system state in the future. Therefore, the selected leaf node embodies the best pose of the robot in the subsequent $T=3$ seconds. Consequently, we designate this pose as the robot's navigational goal point for the current time-step.

\section{RESULTS AND EXPERIMENTS}
\begin{figure}[t]
    \centering
    \includegraphics{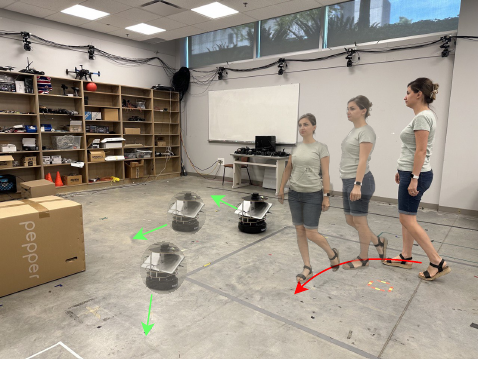}
    \caption{An illustration depicting a real-world experiment. The future trajectory of the human is represented by the red arrow, while the future positions of the robot are displayed, with its orientation indicated by the green arrows.}
    \label{fig:exp_fig}
\end{figure}

To evaluate the performance of the proposed method, three distinct experiments were conducted in both simulation and real-world environments.
A photograph of the real-world experiment is shown in Fig. \ref{fig:general} (left) and Fig. \ref{fig:exp_fig}.
A collection of video demonstrations showcasing all three real-world experiments is available on YouTube\footnote{https://www.youtube.com/playlist?list=PLMSLU9mHe0DlH-nXzEEjmA35DtHhZh5GL}.

\subsection{MCTS-DRL vs MCTS and DRL}

\begin{figure*}
    \centering
    \includegraphics[scale=0.88]{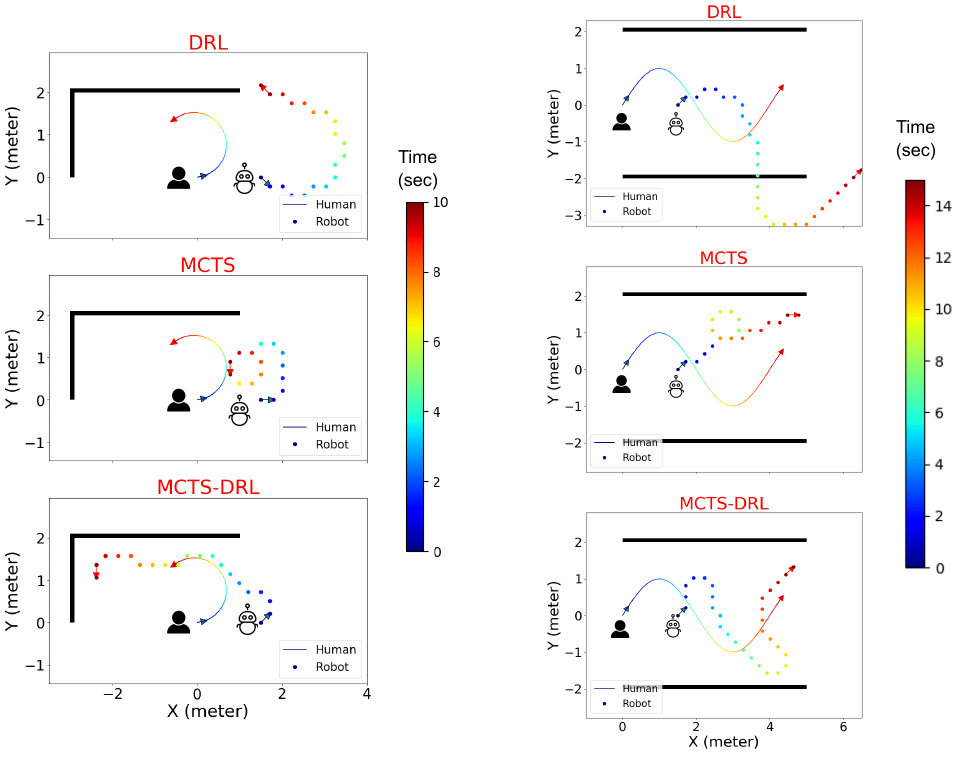}
    \caption{Performance comparison of DRL, MCTS, MCTS-DRL for two different human trajectories. The human and robot trajectories are depicted by a line and points, respectively. The rainbow color scale indicates the time dimension, with purple and red denoting the first and last time-steps, respectively. The MCTS-DRL method outperforms the standalone MCTS and DRL algorithms, effectively following in front of the human, avoiding obstacles, and producing a consistent and stable behavior.}
    \label{fig:exp1}
\end{figure*}
In this section, we compare the performance of the proposed MCTS-DRL method against standard MCTS and DRL algorithms. The experiment was conducted in the simulated environment, and the results are illustrated in Fig. \ref{fig:exp1} for two distinct human trajectories: a circular path and an $S$-shaped path.
The human's trajectory is depicted by a line, while the robot's trajectory is represented by points. The time dimension is indicated by the rainbow color scale, with purple and red denoting the first and last time-steps, respectively.
The results reveal that the DRL approach is unable to effectively follow the human and avoid obstacles, which is attributed to the training of the model in an obstacle-free environment; 
on the other hand, policies trained in environments with obstacles would have a hard time generalizing to other environments with different obstacle configurations.
Also, the MCTS approach fails to generate consistent results, particularly around corners which is due to its random action selection.
In contrast, the proposed MCTS-DRL method demonstrates a superior performance, generating a trajectory for the robot that effectively maintains a specified distance from the human, avoids occlusion and collisions with obstacles, and results in a consistent and stable behavior.

\begin{table}[b]
    \centering
    \caption{Sum of rewards achieved by DRL, MCTS, and MCTS-DRL methods for two distinct human trajectories.}
    \begin{tabular}{cccc}
         Human Trajectory & DRL & MCTS & MCTS-DRL \\ \hline \vspace{0mm}
         &&&\\
         Circle & $-17.95$ & $2.87 \pm 5.96$   &   $\mathbf{4.53}$ \\
         $S$-shaped & $-21.84$ & $-3.83 \pm 4.33$ & $\mathbf{-1.61}$\\
    \end{tabular}
    \label{tab:exp1-rewards}
\end{table}

\begin{table*}
  \centering
  \caption{Follow-ahead comparative results for three human trajectories in an obstacle free environment. The proximity of the distance to 1.5 and the orientation to 0 indicates better performance.}
  \label{tab:obt_free}
  \begin{tabular}{ c  l c c  || c  l c c}  
    Human & Method & Distance (m)  & orientation($\alpha$ (deg))  & Human & Method & Distance (m)  & orientation($\alpha$ (deg)) \\
    Trajectory& & Mean $\pm$ std& Mean $\pm$ std  & Trajectory& & Mean $\pm$ std& Mean $\pm$ std\\\hline

    \multirow{6}{*}{\begin{minipage} {.03\linewidth}
      \includegraphics[scale=0.17]{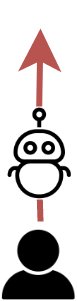}
    \end{minipage}} & & & & \multirow{6}{*}{\begin{minipage}{.03\linewidth}
      \includegraphics[scale=0.17]{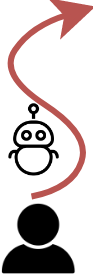}
    \end{minipage}} &&&\\  &&&&&&\\
    & MCTS-DRL & \textbf{1.32 $\pm$ 0.11} &-3.47 $\pm$ 7.8 & & MCTS-DRL & \textbf{1.33 $\pm$ 0.34} & -28.33 $\pm$ 62.24\\
    &LBGP & 1.24 $\pm$ 0.3& \textbf{2.1 $\pm$ 14.5} & &LBGP & 1.86 $\pm$ 0.4 & \textbf{16.9 $\pm$ 28.3}\\  & &&&&& \\  &&&&&&\\
    \hline
    \multirow{6}{*}{\begin{minipage}{.03\linewidth}
      \includegraphics[scale=0.17]{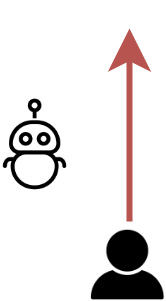}
    \end{minipage}} & &&& \multirow{6}{*}{\begin{minipage}{.03\linewidth}
      \includegraphics[scale=0.17]{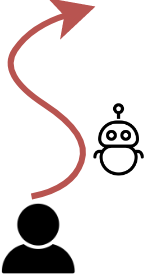}
    \end{minipage}} & &&\\  &&&&&&\\
    & MCTS-DRL & 1.16 $\pm$ 0.18 & \textbf{-15 $\pm$ 16.25} && MCTS-DRL & \textbf{1.20 $\pm$ 0.32} & -19.01 $\pm$ 49.51\\
    &LBGP & \textbf{1.61 $\pm$ 0.4} &30.2 $\pm$ 34.6 &&LBGP & 2.09 $\pm$ 0.3 & \textbf{-4.9 $\pm$ 26.7}\\  & &&&&& \\  &&&&&&\\
    \hline
    \multirow{6}{*}{\begin{minipage}{.03\linewidth}
      \includegraphics[scale=0.17]{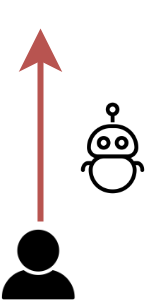}
    \end{minipage}} & &&&\multirow{6}{*}{\begin{minipage}{.03\linewidth}
      \includegraphics[scale=0.17]{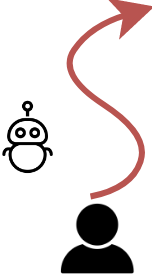}
    \end{minipage}} & && \\  &&&&&&\\
    & MCTS-DRL & 1.12 $\pm$ 0.1 & \textbf{4.7 $\pm$ 17}&& MCTS-DRL & \textbf{1.21 $\pm$ 0.36} & \textbf{-28.64 $\pm$ 55.17} \\
    &LBGP & \textbf{1.63 $\pm$ 0.4} &  -8.4 $\pm$ 26.6 & &LBGP & 1.81 $\pm$ 0.6 & 35.8 $\pm$ 33.1\\  & &&&&& \\  &&&&&&\\
    \hline
    
    \multirow{6}{*}{\begin{minipage}{.03\linewidth}
      \includegraphics[scale=0.17]{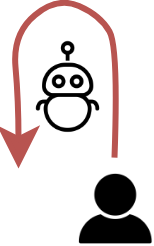}
    \end{minipage}} & &&&\multirow{6}{*}{\begin{minipage}{.03\linewidth}
      \includegraphics[scale=0.17]{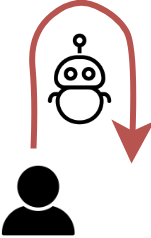}
    \end{minipage}} & && \\  &&&&&&\\
    & MCTS-DRL & \textbf{1.47 $\pm$ 0.32} & \textbf{-8.76 $\pm$ 132.04} && MCTS-DRL & \textbf{1.51 $\pm$ 0.2} & \textbf{12.01 $\pm$ 142.07}\\
    &LBGP & 1.34 $\pm$ 0.4 & 20.9 $\pm$  37.8 & &LBGP & n/a & n/a\\  &&&&&& \\  &&&&&&\\
    \hline

    \multirow{6}{*}{\begin{minipage}{.03\linewidth}
      \includegraphics[scale=0.17]{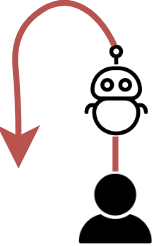}
    \end{minipage}} & &&&\multirow{6}{*}{\begin{minipage}{.03\linewidth}
      \includegraphics[scale=0.17]{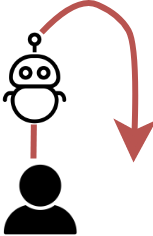}
    \end{minipage}} & && \\  &&&&&&\\
    & MCTS-DRL & \textbf{1.44 $\pm$ 0.33} & 14.39 $\pm$ 119.68&& MCTS-DRL & \textbf{1.53 $\pm$ 0.34} & \textbf{-10.01 $\pm$ 98.97} \\
    &LBGP & 1.99 $\pm$ 0.2  & \textbf{-16.8 $\pm$ 18.6} &&LBGP & n/a &n/a\\  &&&& && \\  &&&&&&\\
    \hline

  \end{tabular}
\end{table*}

Since MCTS generates different trajectories for the robot resulting different return, we conducted 20 experimental trials. The results of the total return achieved for each trajectory are presented in Table \ref{tab:exp1-rewards}. The rewards for all three approaches were calculated using (\ref{eq:r}) at each time-step and the cumulative rewards for each trajectory are shown in the table. The results indicate that the proposed MCTS-DRL algorithm outperforms both the DRL algorithm and the mean value of the MCTS algorithm, as evident from the higher total rewards achieved.

\subsection{Follow-ahead in an obstacle-free environment}

We conducted a comparison between the performance of human following in an obstacle-free environment using the proposed MCTS-DRL method and the LBGP method proposed in \cite{c8}. 
Three distinct trajectories were used in the experiment, including a straight line, an $S$-shaped, and a $U$-shaped path. The target person walked at a linear velocity of 0.6 m/s in the straight path and 0.3 m/s in the curved paths, with an angular velocity of 0.3 rad/s in all cases. The robot's linear and angular velocities were set to  $[0.5, 0.65]$ m/s and $[-2, 2]$ rad/s, respectively, and its initial position relative to the human was varied.

In all experiments, we utilized a motion capture system to track the positions of both the robot and human at each time-step, just for the purpose of calculating the reward value and evaluating the performance of the algorithm. It should be noted that the algorithm only utilized camera depth data to obtain the human pose and the robot's encoder to determine its own pose, and the recorded data was only used for evaluation purposes.
The outcomes of the experiments, in terms of the mean human-robot relative distance and mean orientation angle ($\alpha$), are presented in Table \ref{tab:obt_free}.

The results demonstrated that the robot successfully maintained the human-robot relative distance within the range of $[1, 2]$ m for all trajectories, and attempted to follow the human in front. This experiment demonstrates the efficacy of the proposed MCTS-DRL approach in achieving comparable results to previous methods in obstacle-free environments.

\subsection{Obstacle and occlusion avoidance}
\subsubsection{Performance Evaluation with and without Obstacles}

\begin{figure}[t]
    \centering
    \includegraphics[width=\columnwidth]{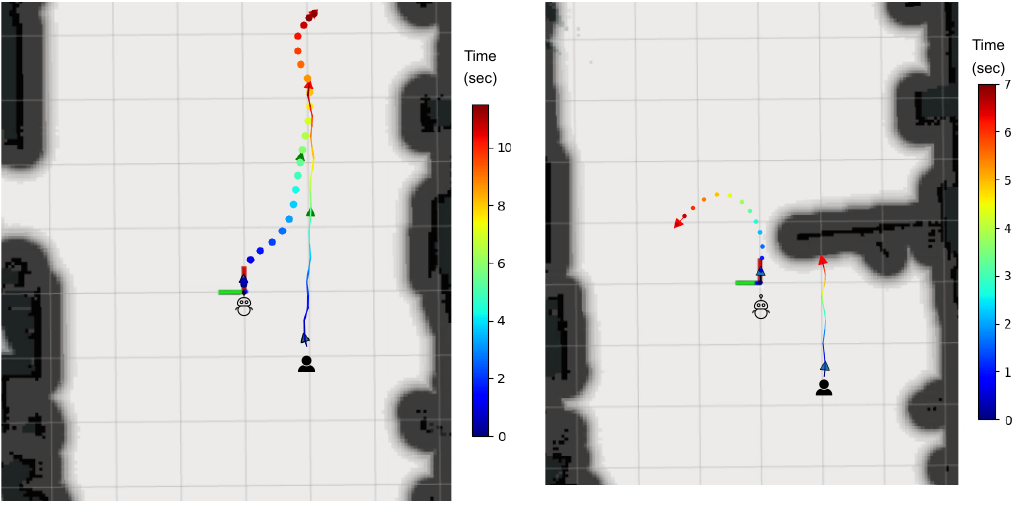}
    \caption{Performance evaluation of MCTS-DRL in the absence of obstacles with the human walking in straight line (left figure) and the presence of an obstacle (right figure) in which the robot turned left to avoid occlusion. The robot and human's trajectories are shown with points and line, respectively. The side bar shows the time starting with the purple and ending with red color, respectively.}
    \label{fig:exp3_straight}
\end{figure}

\begin{figure}
    \centering
    \includegraphics[width=\columnwidth]{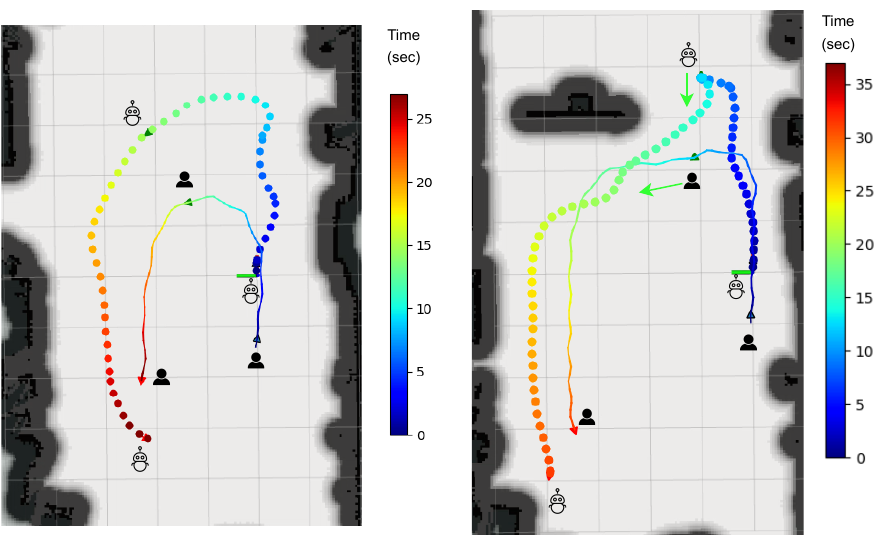}
    \caption{Performance evaluation of MCTS-DRL in the absence of obstacles with the human walking in a $U$-shaped path (left figure) and the presence of an obstacle (right figure) in which the robot changed its direction to avoid occlusion. The robot and human's orientation is shown with the green arrows at the time of direction altering. The robot and human's trajectories are shown with points and line, respectively. The side bar shows the time starting with the purple and ending with red color, respectively.}
    \label{fig:exp3_u}
\end{figure}

\begin{figure}
    \centering
    \includegraphics[width=\columnwidth]{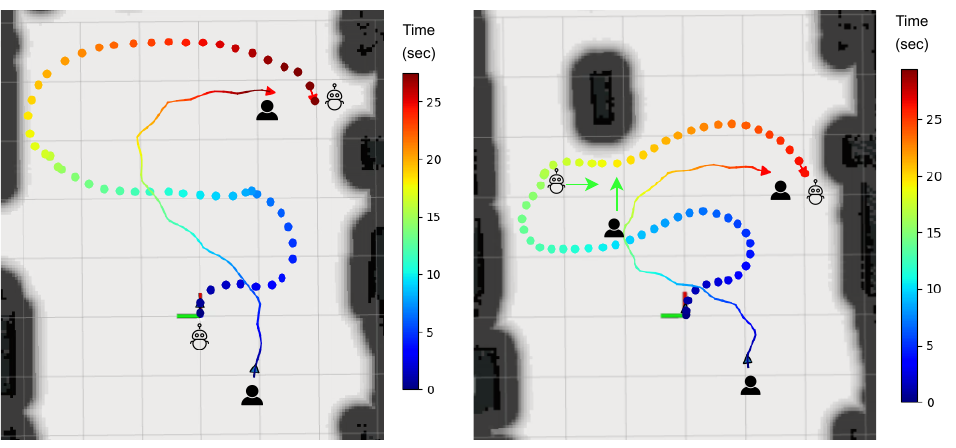}
    \caption{Performance evaluation of MCTS-DRL in the absence of obstacles with the human walking in a $S$-shaped path (left figure) and the presence of an obstacle (right figure) in which the robot changed its direction to avoid occlusion. The robot and human's orientation is shown with the green arrows at the time of direction altering. The robot and human's trajectories are shown with points and line, respectively. The side bar shows the time starting with the purple and ending with red color, respectively.}
    \label{fig:exp3_s}
\end{figure}

Since there are no existing follow-ahead methods proposed for environments with obstacles, comparisons with other methods were not possible. Therefore, the following scenarios were conducted to evaluate the performance of the proposed method in the presence and absence of obstacles within the environment. 
Our algorithm is capable of following the target person along any random trajectory. However, in order to ensure consistency with previous studies, we evaluated its performance with three distinct trajectories: straight, $U$-shaped, and $S$-shaped.
As with the previous experiments, the positions of both the human and robot were tracked using the motion capture system for evaluation purposes.
For all three experiments, we compared the performance of the robot with and without obstacles present in the environment. As shown in Fig. \ref{fig:exp3_straight}, the robot followed in front of the person with no obstacles in the vicinity. However, when the robot approached an obstacle, it turned left to avoid occlusion. In Fig. \ref{fig:exp3_u}, the $U$-shaped trajectory is illustrated. With the presence of an obstacle, the robot altered its path at a time of 12 s to avoid occlusion rather than rotating around the obstacle. The $S$-shaped trajectory is depicted in Fig. \ref{fig:exp3_s}. Similar to the $U$-shaped trajectory, the robot changed its path at 17 s to avoid occlusion. The orientation of the robot and human during the $U$-shaped and $S$-shaped trajectories is represented by the green arrows at the time of directional changes for occlusion avoidance purposes.

\subsubsection{Performance Evaluation in a $L$-shaped environment}

\begin{figure}[t]
    \centering
    \includegraphics[width=0.8\columnwidth]{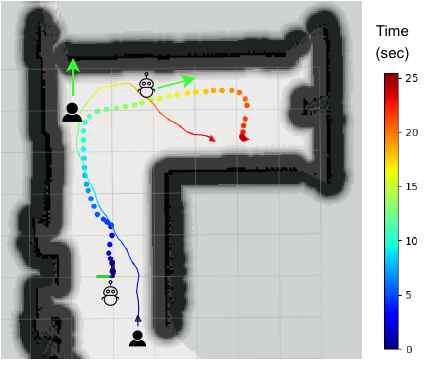}
    \caption{Trajectory of the robot and human navigating an $L\_$shaped path. The trajectory of the human and robot are shown with line and points, respectively. The green arrows indicate the orientation of the robot and human when the human walks toward the corner. The robot avoids colliding with obstacles and stays beside the human subject, navigating ahead of them subsequently.}
    \label{fig:exp3-L}
\end{figure}

We conducted an evaluation of the performance of the proposed MCTS-DRL method in navigating corridors, specifically an $L$-shaped path. The trajectory of both the human and robot was recorded, and the results are presented in Fig. \ref{fig:exp3-L}. As illustrated, the human walked towards the corner of the corridor, while the robot, to avoid colliding with obstacles, turned right and remained beside the human. The green arrows indicate the orientation of the robot and human at this juncture. Subsequently, the robot continued to navigate ahead of the human successfully.

\section{conclusion}

This study introduces a novel methodology for robotic follow-ahead applications in avoiding collision and occlusions caused by the surrounding obstacles. By integrating MCTS with DRL, our approach generates reliable navigational goals for the robot. Through three different experiments, we demonstrated the superiority of our proposed MCTS-DRL approach over pure MCTS and DRL algorithms, as it effectively follows a target person in front of him/her, maintaining a defined distance in the presence and absence of obstacles in the environment. These results highlight the potential of our MCTS-DRL approach to enhance autonomous robotic navigation and overcome collision and occlusion challenges in real-world environments.

% it was 12
\addtolength{\textheight}{-15cm}   % This command serves to balance the column lengths

%%%%%%%%%%%%%%%%%%%%%%%%%%%%%%%%%%%%%%%%%%%%%%%%%%%%%%%%%%%%%%%%%%%%%%%%%%%%%%%%

%%%%%%%%%%%%%%%%%%%%%%%%%%%%%%%%%%%%%%%%%%%%%%%%%%%%%%%%%%%%%%%%%%%%%%%%%%%%%%%%

%%%%%%%%%%%%%%%%%%%%%%%%%%%%%%%%%%%%%%%%%%%%%%%%%%%%%%%%%%%%%%%%%%%%%%%%%%%%%%%%

\bibliographystyle{IEEEtran}
\bibliography{ref}

\end{document}